\documentclass[sigconf]{acmart}
\usepackage{graphicx}
\usepackage{subfigure}

\begin{document}


%

\title{Multi-modal Conditional Attention Fusion for Dimensional Emotion Prediction}

%
%
%
%
%

%
\author{Shizhe Chen}
\affiliation{%
  \institution{Multimedia Computing Lab\\School of Information\\Renmin University of China}
}
\email{cszhe1@ruc.edu.cn}

\author{Qin Jin}
\authornote{Corresponding author.}
\affiliation{%
  \institution{Multimedia Computing Lab\\School of Information\\Renmin University of China}
}
\email{qjin@ruc.edu.cn}

\begin{abstract}
Continuous dimensional emotion prediction is a challenging task where the fusion of various modalities usually achieves state-of-the-art performance such as early fusion or late fusion. In this paper, we propose a novel multi-modal fusion strategy named conditional attention fusion, which can dynamically pay attention to different modalities at each time step. Long-short term memory recurrent neural networks (LSTM-RNN) is applied as the basic uni-modality model to capture long time dependencies. The weights assigned to different modalities are automatically decided by the current input features and recent history information rather than being fixed at any kinds of situation. Our experimental results on a benchmark dataset AVEC2015 show the effectiveness of our method which outperforms several common fusion strategies for valence prediction.
\end{abstract}

%
%

%
%

%
%



\maketitle

\section{Introduction}
Understanding human emotions is a key component to improve human-computer interactions \cite{Picard2003Affective}.
A wide range of applications can benefit from emotion recognition such as customer call center, computer tutoring systems and mental health diagnoses.\par

Dimensional emotion is one of the most popular computing models for emotion recognition \cite{marsella2014computationally}.
It maps an emotion state into a point in a continuous space.
Typically the space consists of three dimensions: arousal (a measure of affective activation), valence (a measure of pleasure) and dominance (a measure of power or control).
This representation can express natural, subtle and complicated emotions.
There have been many research works on dimensional emotion analysis for better understanding human emotions in recent years \cite{wollmer2013lstm, valstar2014avec, Ringeval2015AV}.\par

Since emotions are conveyed through various human behaviours, past works have utilized a broad range of modalities for emotion recognition including speech \cite{el2011survey}, text \cite{socher2013recursive}, facial expression \cite{fasel2003automatic}, gesture \cite{piana2014real}, physiological signals \cite{DBLP:journals/ijon/WangNL14}, etc. Among them, facial expression and speech are the most common channels to transmit human emotions. It is beneficial to use multiple modalities for emotion recognition.\par

Fusion strategies for different modalities in previous works can be divided into 3 categories, namely feature-level (early) fusion, decision-level (late) fusion and model-level fusion \cite{wu2014survey}. 
Early fusion uses the concatenated features from different modalities as input features for classifiers. It has been widely used in the literature to successfully improve performance \cite{DBLP:conf/interspeech/RozgicASKVP12}. However, it suffers from the curse of dimensionality. Also it's not very useful when features are not synchronized in time. 
Late fusion eliminates some disadvantages of early fusion. It combines the predictions of different modalities and trains a second level model such as RVM \cite{Huang2015An}, BLSTM \cite{DBLP:conf/mm/HeJYPWS15}. But it ignores interactions and correlations between different modality features. 
Model-level fusion is a compromise between the two extremes. The implementation of model-level fusion depends on the specific classifiers. For example, for neural networks, model-level fusion could be concatenation of different hidden layers from different modalities \cite{DBLP:conf/mm/WuJWPX14}. For kernel classifiers, model-level fusion could be kernel fusion \cite{DBLP:conf/icmi/ChenCCF14}. As for Hidden Markov Model (HMM) classifiers, novel forms of feature interactions have been proposed \cite{DBLP:conf/icpr/LuJ12}.\par

\begin{figure*}
\center{\includegraphics[width=0.85\linewidth]{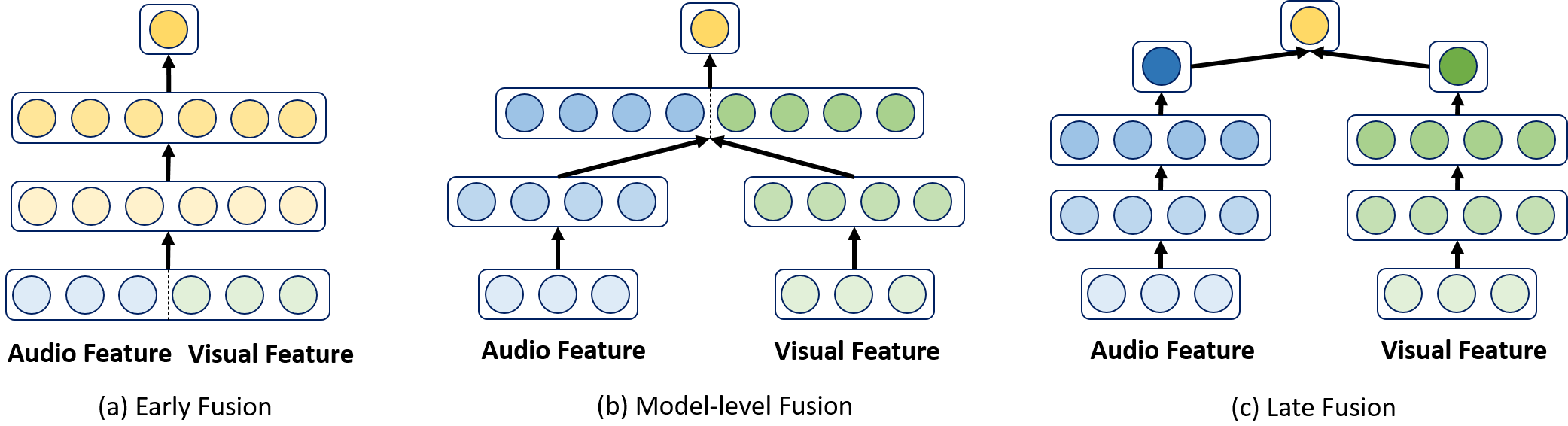}}
\caption{Three typical multi-modality fusion strategies. (a) Early fusion: concatenation of features from different modalities. (b) Model-level fusion: concatenation of high level feature representations from different modalities. (c) Late fusion: fusion of predictions from different modalities.}
\label{fig:typical_fusion_structures}
\end{figure*}

\begin{figure}
\center{\includegraphics[width=0.6\linewidth]{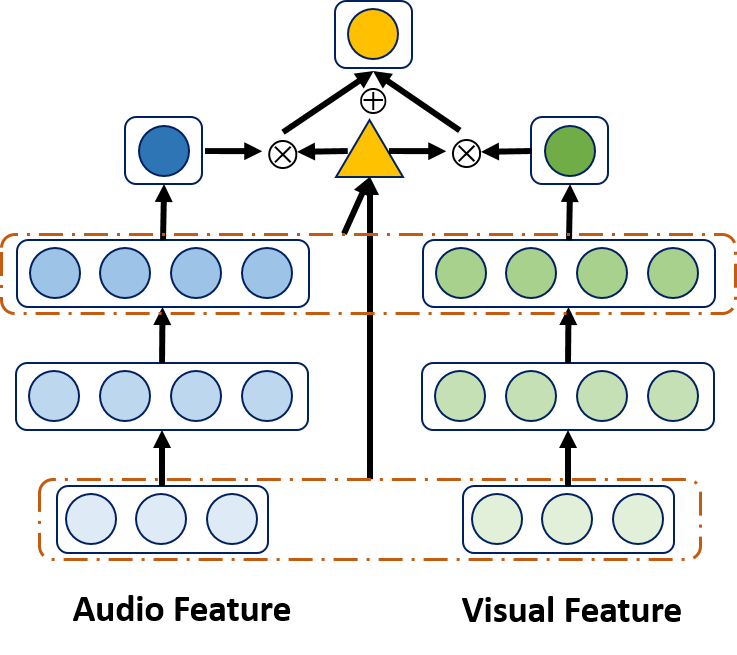}}
\caption{Conditional attention fusion model. The units in the dashed red line is concatenated together as the input of the triangle to learn the attention weights. $\otimes$ and $\oplus$ denote the element multiplication and addition respectively.}
\label{fig:ca_structure}
\end{figure}

In this paper, we propose a novel architecture for the fusion of different modalities called conditional attention fusion. 
We use Long-short term memory recurrent neural networks (LSTMs) as the basic model for each uni-modality since LSTMs are able to capture long time dependencies. For each time step, the fusion model learns how much of attentions it should put on each modality conditioning on its current input multi-modal features and recent history information. This approach is similar to human perceptions since humans can dynamically focus on more obvious and trustful modalities to understand emotions.
Unlike early fusion, we dynamically combine predictions of different modalities, which avoids the curse of dimensionality and synchronization between different features. And unlike late fusion, the input features are interacted in a higher level to learn the current attention instead of being isolated without any interactions among different modalities. The main architecture is shown in Figure~\ref{fig:ca_structure}. We use the AVEC2015 dimensional emotion dataset \cite{Ringeval2015AV} to evaluate our methods. The results shows the effectiveness of our new fusion architecture.\par

\section{Multi-modal Features}
\subsection{Audio Features}
We utilize the OpenSMILE toolkit \cite{Eyben2010Opensmile} to extract low-level features including MFCCs, loudness, F0, jitter and shimmer. All the features are extracted using 40ms frame window size without overlap to match with the groundtruth labels since it is demonstrated in \cite{DBLP:conf/mm/ChenJ15} that short-time features can reveal more details  and thus boost performance for affective prediction using LSTMs. The low-level acoustic features are in 76 dimensions.

\subsection{Visual Features}
Two sets of visual features are extracted from facial expression: appearance-based features and geometric-based features \cite{Ringeval2015AV}. The appearance-based features are computed by using Local Gabor Binary Patterns from Three Orthogonal Planes (LGBP-TOP) and are compressed to 84 dimensions via PCA. The geometric-based features in 316 dimensions are computed from 49 facial landmarks. Frames where no face is detected are filled with zeros. We concatenate appearance-based features and geometric-based features as our visual feature representations.
\\
\section{Emotion Prediction Model}
\subsection{Uni-Modality Prediction Model}
Long short term memory (LSTM) architecture \cite{DBLP:journals/neco/HochreiterS97} is the state-of-the-art model for sequence analysis and can exploit long range dependencies in the data. In this paper, we use the peephole LSTM version proposed by Graves \cite{Graves2013Generating}. The function of hidden cells and gates are defined as follows.
\begin{gather}
i_{t} = \sigma (W_{xi}x_{t} + W_{hi}h_{t-1} + W_{ci}c_{t-1} + b_{i})\nonumber\\
f_{t} = \sigma (W_{xf}x_{t} + W_{hf}h_{t-1} + W_{cf}c_{t-1} + b_{f})\nonumber\\
c_{t} = f_{t} \cdot c_{t-1} + i_{t} \cdot tanh(W_{xc}x_{t} + W_{hc}h_{t-1} + b_{c})\\
o_{t} = \sigma (W_{xo}x_{t} + W_{ho}h_{t-1} + W_{co}c_{t-1} + b_{o})\nonumber\\
h_{t} = o_{t} \cdot tanh(c_{t})\nonumber
\label{eqn:lstm_cell}
\end{gather}

where $i, f, o$ and $c$ refers to the input gate, forget gate, output gate, and cell state respectively. $\sigma(\cdot)$ is the sigmoid function and $tanh(\cdot)$ is the tangent function.

\subsection{Conditional Attention Fusion Model}
Let $x_{t}^{a}$ and $x_{t}^{v}$ refer to the audio features and visual features respectively at the $t^{th}$ frame. $h_{t}^{a}$ and $h_{t}^{v}$ refer to the outputs of the last hidden layer from uni-modality model with audio or visual features respectively. $f_{\theta_{a}}$ and $f_{\theta_{v}}$ refer to the uni-modality model which maps the audio or visual features into predictions. We define the conditional attention fusion of the predictions from the two modalities at timestep $t$ as: 
\begin{gather}
\widehat{y_{t}} = \lambda_{t} \cdot f_{\theta_{a}}(x_{t}^{a},h_{t-1}^{a}) + (1-\lambda_{t}) \cdot  f_{\theta_{v}}(x_{t}^{v},h_{t-1}^{v})\\
\lambda_{t} = \sigma(W_{g}[h_{t}^{a}|h_{t}^{v}|x_{t}^{a}|x_{t}^{v}])
\label{eqn:conditional_attention}
\end{gather}
where $[h_{t}^{a}|h_{t}^{v}|x_{t}^{a}|x_{t}^{v}]$ is the concatenation of the representations inside the bracket.\par
The $\lambda_{t}$ is calculated based on the current audio and visual features and their high-level history information for two reasons. Firstly, the current input features are the most direct indicators to show whether the modality is reliable. For example, for facial features, inputs filled with 0s suggest that the current face detection fails and thus should be assigned with less confidence. Secondly, the weights assigned to each modality would be smoothed by considering high-level history features $h_{t}^{v}$ and $h_{t}^{a}$ in addition to current input features. In this way, the model can dynamically pay attention to different modalities, which could improve the stability in different situations.

\subsection{Model Learning}
Intuitively, the acoustic features are more reliable when the acoustic energy is higher, because the headset microphone can record speech from both the subject speaker and other speakers in conversations. Higher energy may refers to higher confidence that the speech is from the target subject. Similarly the facial features are reliable only when faces are correctly detected. So adding such side information might be beneficial to learn the attention weights.\par
We transform the acoustic energy into scale [0, 1], and we use $g_{t}^{a}$ to denote its value at the $t^{th}$ timestep. For visual features, we use $g_{t}^{v} \in \{0, 1\}$ to indicate whether the subject's face is detected since the face detection provided in the dataset has no detection confidence. We therefore define the final loss function for one sequence as follows:
\begin{gather}
L_{t}^{g} = \frac{1}{2}(\alpha (g_{t}^{a}-\lambda_{t})^{2}+\beta (g_{t}^{v}-(1-\lambda_{t}))^{2})\\
L = \sum_{t}\frac{1}{2}(\widehat{y_{t}}-y_{t})^{2}+L_{t}^{g}
\label{eqn:ca_loss}
\end{gather}
where $\alpha$ and $\beta$ are hyper-parameters and are optimized on the development set. In practice, $\alpha$ and $\beta$ are usually set to small values around $10^{-2}$ to avoid $L_{t}^{g}$ over-affecting on $\lambda_{t}$.\par
The derivative of $L_{t}^{g}$ with respect to $\lambda_{t}$ is as follows:
\begin{gather}
\frac{\partial L_{t}^{g}}{\partial \lambda_{t}} = \beta g_{t}^{v}-\alpha g_{t}^{a} - \beta + (\alpha + \beta)\lambda_{t}
\label{eqn:loss_gate_derivative}
\end{gather}
When $g_{t}^{a}$ is high and $g_{t}^{v}$ is low, (\ref{eqn:loss_gate_derivative}) is close to $(\alpha+\beta)(\lambda_{t}-1)$ (the extreme case when $g_{t}^{a}=1$ and $g_{t}^{v}=0$). The derivative is less than 0, which will push $\lambda_{t}$ to increase to give acoustic features more confidence. But when $g_{t}^{a} \to 0$ and $g_{t}^{v} \to 1$, (\ref{eqn:loss_gate_derivative}) is close to $(\alpha + \beta)\lambda_{t}$, which is larger than 0 and will push $\lambda_{t}$ to decrease to focus on visual features. When $g_{t}^{a} \approx g_{t}^{v}$, the absolute value of the derivative would be small and thus $L_{t}^{g}$ has less influence on $\lambda_{t}$.
\section{Experiments}
\subsection{Dataset}
The AVEC2015 dimensional emotion dataset is a subset of the RECOLA dataset \cite{DBLP:conf/fgr/RingevalSSL13}, a multimodal corpus of remote and collaborative affective interactions. There are 27 subjects in the dataset and are equally divided into training, development and testing sets. Audio, video and physiological data are collected for each participant for the first 5 minutes of interactions. Arousal and valence are annotated in scale [-1, 1] for every 40ms \cite{Ringeval2015AV}. Since the submission times on testing set are limited, we carry out cross validation on the development set. We randomly select 5 subjects as the development set to optimize hyper parameters and the remained 4 speakers are used as the test set. We do the experiments 8 times. The concordance correlation coefficient (CCC) \cite{Ringeval2015AV} works as the evaluation metric.

\subsection{Experimental Setup}
Annotation delay compensation \cite{Huang2015An} is applied because there exists a delay between signal content and groundtruth labels due to annotators' perceptual processing. We drop first $N$ groundtruth labels and last $N$ input feature frames. $N$ is optimized by non-temporal regression model SVR on training set. In this paper, $N$ is optimized to be 20 frames for both audio and visual features. When predicting the result, the outputs of the model are shifted back by $N$ frames. The missing predictions in the first $N$ frames are filled with zeros. Finally, a binomial filter is applied to smooth the predictions. Annotation delay compensation and smoothing is applied in all the following experiments.\par
The input features are normalized into the range [-1,1]. For acoustic features, the LSTM has 2 layers and 100 cells for each layer. For visual features, the LSTM has 2 layers and 120 cells for each layer. The size of mini-batch is 256 and truncated backpropagation through time (BPTT) \cite{Werbos1990Backpropagation} is applied. The initial learning rate is set to be 0.01 with learning rate decay. Dropout is used as regularization. The training epochs are 100 and the model that achieves the best performance in development set is used as the final model.\par
We compare the conditional attention fusion model with early fusion, late fusion and model-level fusion. For early fusion, the LSTM has 150 units each layer, which has the similar size of parameters to other fusion methods. For late, model-level and conditional attention fusion, the parameters in LSTM are initialized with the trained uni-modal LSTMs. In order to avoid overfitting, we only fine-tune the network for 10 epochs with smaller initial learning rate 0.001. The hyper-parameters $\alpha$ and $\beta$ are set to zeros for arousal prediction and 0.04, 0.02 respectively for valence prediction.

\begin{table}
\centering
\caption{CCC performance of uni-modal features}
\begin{tabular}{ccc} \hline
&audio feature & visual feature\\ \hline
arousal & 0.787 & 0.432\\
valence & 0.595 & 0.620\\ \hline
\end{tabular}
\label{tab:unimodal_results}
\end{table}

\begin{table}
\centering
\caption{CCC performance of different loss functions on valence prediction}
\begin{tabular}{ccc} \hline
&mean & std\\ \hline
without $L_{t}^{g}$, $\alpha=\beta=0$ & 0.672 & 0.046\\
with $L_{t}^{g}$, $\alpha=0.04, \beta=0.02$ & 0.684 & 0.041\\ \hline
\end{tabular}
\label{tab:alpha_beta_results}
\end{table}

\begin{figure}
\center{\includegraphics[width=0.8\linewidth]{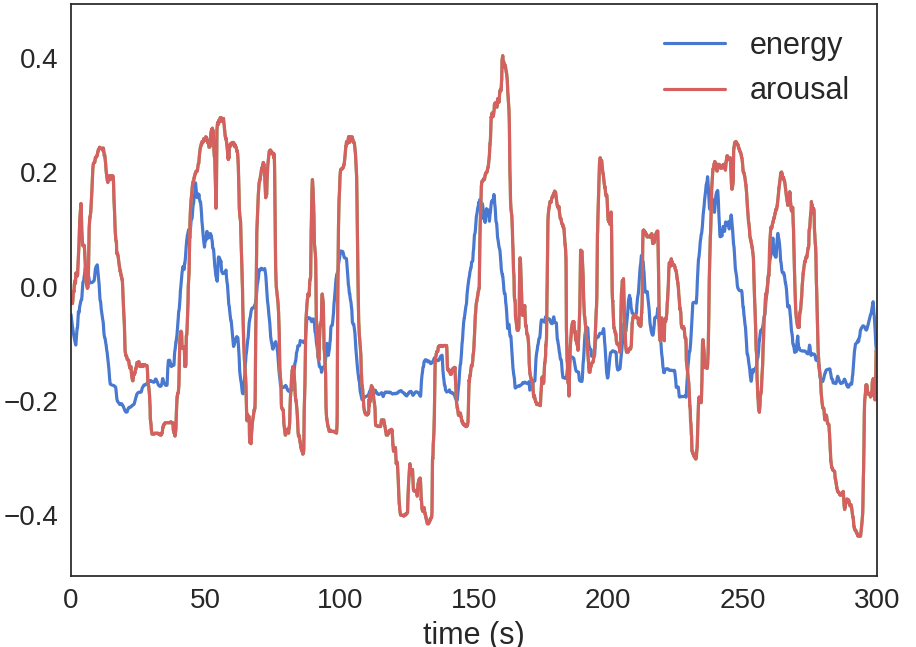}}
\caption{The relationship between arousal labels and acoustic energy on an example from dev set}
\label{fig:arousal_energy_relation}
\end{figure}

\subsection{Experimental Results}
Table~\ref{tab:unimodal_results} shows the prediction performance using uni-modality features. Acoustic features achieve the best performance on the arousal prediction and visual features are slightly better than acoustic features on valence prediction.\par


\begin{figure}
\center{\includegraphics[width=0.75\linewidth]{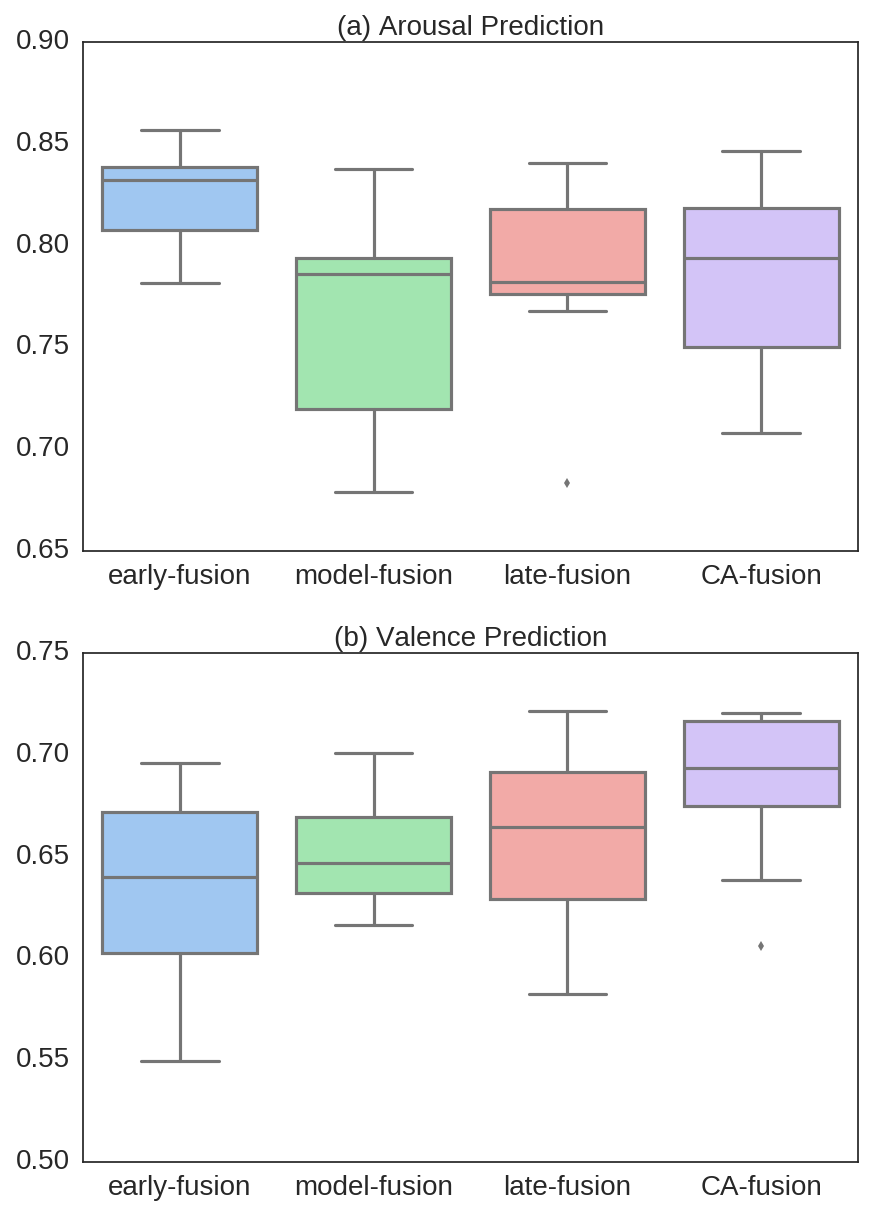}}
\caption{Performance of different fusion strategies}
\label{fig:fusion_results}
\end{figure}

The performance of different fusion methods on arousal prediction is shown in Figure~\ref{fig:fusion_results}(a). Early fusion achieves the average best performance and our proposed fusion method performs the second best among all the fusion strategies. However, there is no significant difference between early fusion prediction and acoustic uni-modality prediction comparing Figure~\ref{fig:fusion_results}(a) with Table~\ref{tab:unimodal_results} (Student t-test with p-value = 0.07). We find that there exists a strong correlation between arousal and the acoustic energy, as shown in Figure~\ref{fig:arousal_energy_relation} where we smooth the acoustic energy with window 100 and shift and scale it according to the mean and standard deviation between energy and arousal labels. And their Pearson Correlation Coefficient on development set is high to 0.558 and CCC is 0.4. This suggests that humans' perception of arousal may mainly base on acoustic features so fusing other modalities may bring less benefit.\par

But for valence prediction, all the fusion strategies outperform the original uni-modality models (as shown in Table 1). An interesting finding is that the higher level the fusion strategy applies, the better performance is achieved as shown in Figure~\ref{fig:fusion_results}(b). Among them, our proposed conditional attention fusion model achieves the best performance and significantly surpasses other fusion strategies by t-test (p<0.02 compared with the second best fusion strategy late fusion, and p<0.007 compared with others). This indicates that dynamically adapting fusion weights for different modalities is beneficial.\par

Table~\ref{tab:alpha_beta_results} shows the CCC performance with and without $L_{t}^{g}$ in loss function. We can see that considering $L_{t}^{g}$ in loss function can further improve performance since it helps to guide the importance of different modalities. It is might because of the insufficiency of data that the model is unable to learn the attention weights effectively without any supervised information. We also observe in our experiments that when $\alpha$ and $\beta$ are around $10^{-2}$, there is no significant change in prediction performance.


\begin{figure}
\center{\includegraphics[width=1\linewidth]{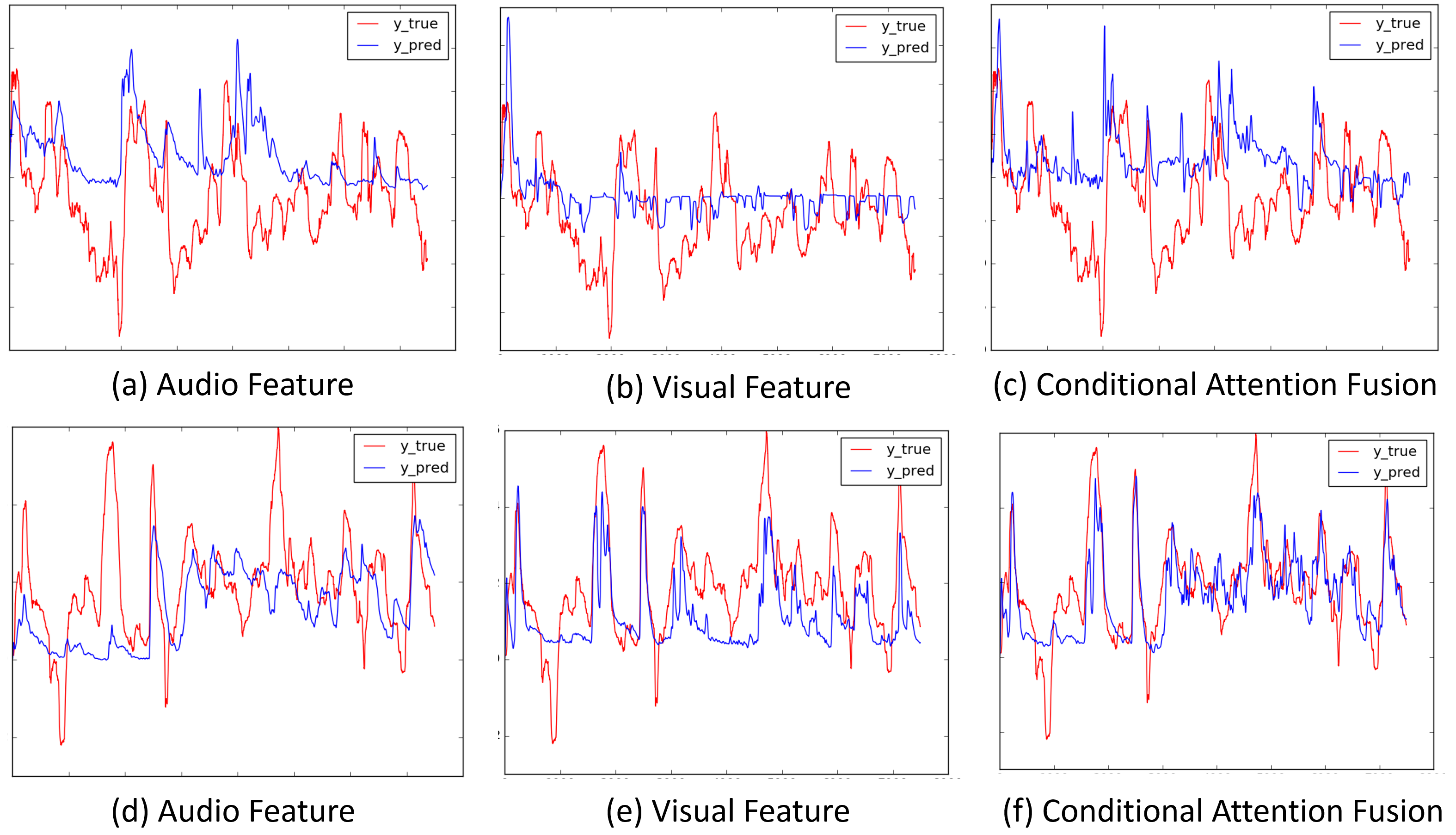}}
\caption{Examples of valence prediction on dev set. The red line is the groundtruth label and the blue line is the prediction.}
\label{fig:fusion_case}
\end{figure}

\begin{table}
\centering
\caption{Valence prediction performance on test set}
\begin{tabular}{cccc} \hline
&rmse & pcc & ccc\\ \hline
Chen et al. \cite{DBLP:conf/mm/ChenJ15} & 0.111 & 0.59 & 0.567\\
Chao et al. \cite{DBLP:conf/mm/ChaoTYLW15} & 0.103 & 0.627 & 0.618 \\ 
CA-fusion & 0.090 & 0.716 & 0.664 \\ \hline
\end{tabular}
\label{tab:test_set}
\end{table}

Figure~\ref{fig:fusion_case} shows some examples of the emotion predictions from the conditional attention fusion method. The upper row shows the case where most of the visual features are missing and the bottom row is another case where the visual features can be extracted in most frames. We can see that the fusion method can make use of the complementary information automatically from audio and visual features in these two situations.

The valence prediction performance of the conditional attention fusion method on testing set is shown in Table ~\ref{tab:test_set}. Chen et al. \cite{DBLP:conf/mm/ChenJ15} use the same feature set as ours and Chao et al. \cite{DBLP:conf/mm/ChaoTYLW15} use more features including CNNs for valence prediction. The comparison further demonstrates the effectiveness of the conditional attention fusion method.
\section{Conclusions}
In this paper we propose a multi-modal fusion strategy named conditional attention fusion for continuous dimensional emotion prediction based on LSTM-RNN. It can dynamically pay attention to different modalities according to the current modality features and history information, which increases the model's stability. Experiments on benchmark dataset AVEC 2015 show that our proposed fusion approach significantly outperform the other common fusion approaches including early fusion, model-level fusion and late fusion for valence prediction. 
In the future, we will use more features from different modalities and apply strategies to express the correlation and independence of different modality features better.


\section{Acknowledgements}
This work is supported by National Key Research and Development Plan under Grant No. 2016YFB1001202.

%
\bibliographystyle{unsrt}
\bibliography{sigproc}  
%
%

\end{document}